\documentclass[letterpaper, 10 pt, conference]{format/ieeeconf}
\usepackage{amsmath}
\usepackage{amsfonts}
\usepackage{setspace}
\usepackage[space, compress, sort]{cite}

\usepackage{amsthm}
\usepackage{amssymb,stmaryrd}
\usepackage[ruled,vlined]{algorithm2e}
\usepackage{mathtools}
\usepackage{tabularx}
\usepackage{graphicx}
\usepackage{subfigure}
\usepackage{enumerate}
\usepackage{float}
\usepackage{url}
\usepackage{verbatim}
\usepackage{booktabs} 
\usepackage{multirow}
\usepackage[linkcolor=black,citecolor=black,urlcolor=black,colorlinks=true]{hyperref}
\usepackage{bm}
\usepackage{empheq}
\IEEEoverridecommandlockouts
\usepackage{graphicx}
\usepackage{siunitx}

\usepackage{textcase}
\usepackage[tablename=TABLE]{caption}
\DeclareCaptionTextFormat{up}{\MakeTextUppercase{#1}}

\title{\LARGE \bf
A Benchmark Comparison of Imitation Learning-based Control Policies for Autonomous Racing
}

\author{Xiatao Sun, Mingyan Zhou, Zhijun Zhuang, Shuo Yang, Johannes Betz, Rahul Mangharam 
\thanks{All authors are with the University of Pennsylvania, Department of Electrical and Systems Engineering, 19104, Philadelphia, PA, USA. Emails: \{\tt\footnotesize sxt, derekzmy, zhijunz, yangs1, joebetz, rahulm\}@seas.upenn.edu}}

\begin{document}

\maketitle
\thispagestyle{empty}
\pagestyle{empty}

\begin{abstract}

Autonomous racing with scaled race cars has gained increasing attention as an effective approach for developing perception, planning and control algorithms for safe autonomous driving at the limits of the vehicle’s handling. To train agile control policies for autonomous racing, learning-based approaches largely utilize reinforcement learning, albeit with mixed results. In this study, we benchmark a variety of imitation learning policies for racing vehicles that are applied directly or for bootstrapping reinforcement learning both in simulation and on scaled real-world environments. We show that interactive imitation learning techniques outperform traditional imitation learning methods and can greatly improve the performance of reinforcement learning policies by bootstrapping thanks to its better sample efficiency. Our benchmarks provide a foundation for future research on autonomous racing using Imitation Learning and Reinforcement Learning.

\end{abstract}

\section{INTRODUCTION}
\subsection{Motivation}
In motorsport racing, it all boils down to the ability of the driver to operate the racecar at its limits. Expert race drivers are extremely proficient in pushing the racecar to its dynamical limits of handling, while accounting for changes in the vehicle's interaction with the environment to overtake competitors at speeds exceeding 300 km/h. Autonomous racing emphasizes driving vehicles autonomously with high performance in racing conditions, which usually involves high speeds, low reaction times, operating at the limits of vehicle dynamics, and constantly balancing safety and performance\cite{betz2022autonomous}. While the goal of autonomous racing is to outperform human drivers through the development of perception, planning and control algorithms, the performance with learning-based approaches is still far from parity. The goal of this paper is to benchmark a variety of imitation learning (IL) approaches that are used directly and for bootstrapping reinforcement learning (RL). 

In the past few years, autonomous racing cars at different scales such as Roborace \cite{Rieber2004ROBORACE}, Indy Autonomous Challenge \cite{Wischnewski2022IndyAuto}, and Formula Student Driverless \cite{Zeilinger2017FSD}, reduced-scale platforms like F1TENTH \cite{okelly2020f1tenth} have been developed. Reduced-scale platforms with on-board computation and assisted with algorithm development in simulation enable rapid development with lower cost for research and educational purposes. 

Autonomous racing has traditionally followed the perception–planning–control modular pipeline. A recent shift towards the end-to-end learning paradigm for autonomous vehicles is showing promise in terms of scaling across common driving scenarios and navigating rare  operation contexts\cite{le2022survey}. Autonomous racing provides a perfect setting for evaluating end-to-end testing approaches as it clearly specifies the trade-off between safety and performance. However, the difficulties in sim-to-real transfer and ensuring safety remain open for further study \cite{betz2022autonomous}. 

\begin{figure}[t]
      \centering
      \includegraphics[scale=0.303]{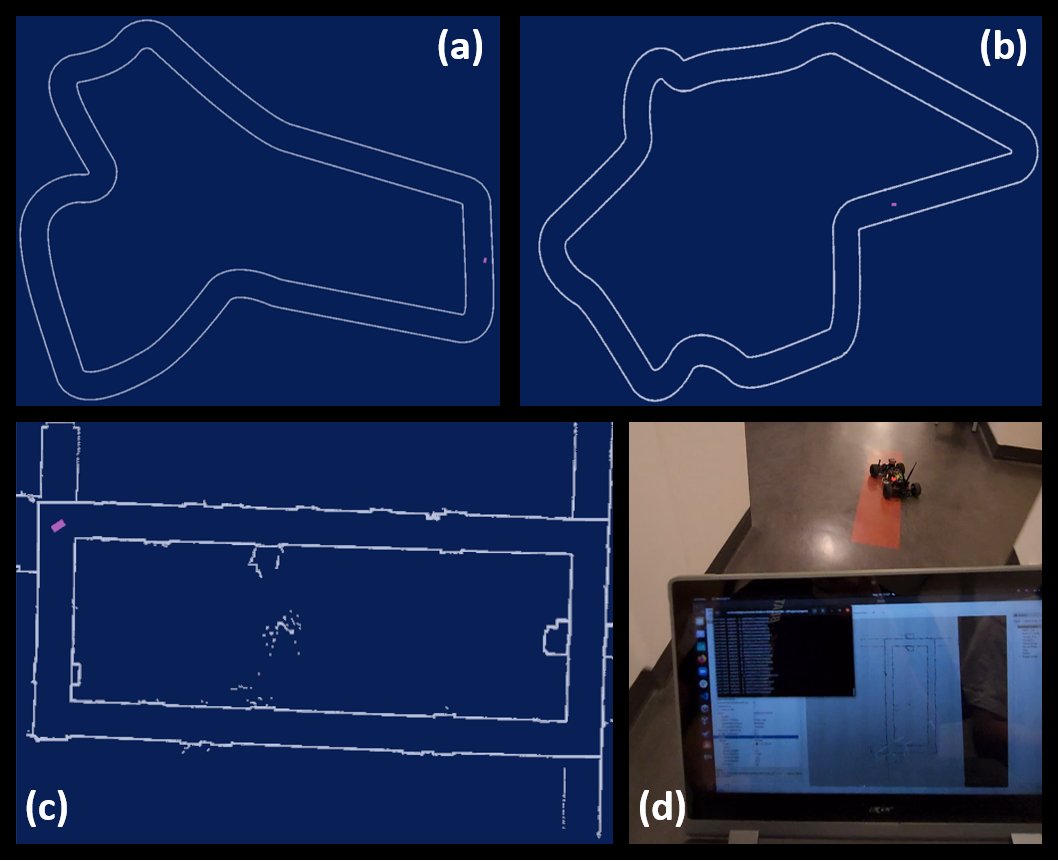}
      \vspace{-5pt}
      \caption{\small{(a) The training map for evaluations and comparisons in simulation. (b) The unseen map for testing the generalizability of learned policies. (c) The training map for the real-world comparison is generated using LiDAR scans of the real environment. (d) F1TENTH vehicle is controlled by the learned policy.}}
      \label{maps}
      \vspace{-20pt}
\end{figure}

Among the emerging end-to-end approaches, IL and RL are the most promising. IL essentially trains policies to mimic the given expert demonstration \cite{ahmed2017survey}. It is shown to outperform supervised machine learning algorithms since those methods suffer from problems including distribution mismatch among datasets and long-term sequential planning \cite{ross2011dagger}. Based on the innovative algorithm Data Aggregation (DAGGER) \cite{ross2011dagger}, some interactive IL methods, such as human-gated DAGGER (HG-DAGGER) \cite{kelly2019hgdagger} and expert intervention learning (EIL) \cite{spencer2022eil}, use interactive querying to improve the training process and overall performance.

However, imitation learning-based autonomous racing vehicles is only just getting started \cite{le2022survey}. Several recent efforts \cite{Pan2018DIL, Cai2021DIRL} implemented IL on autonomous racing cars, but only for bootstrapping and simplified tests. This work implements and provides a comprehensive comparison among several IL methods in simulation and on the F1Tenth  physical racing platform (\url{https://f1tenth.org}). By making this available as open-source software, we hope to encourage researchers to further the exploration in learning-based controllers for autonomous driving.

\subsection{Contributions}
In this paper, we tackle the problem of IL-based control for autonomous ground robots that will drive with high-speed and high acceleration. This work has three main contributions:
\begin{enumerate}

    \item We implement 4 different IL algorithms that learn from expert demonstrations;
    \item We display results from both simulation and real-world experiments on a small-scale autonomous racing car;
    \item We benchmark different algorithms for both direct learning and bootstrapping.
\end{enumerate}

\section{RELATED WORK}

\subsection{Autonomous Racing}
End-to-end approaches for autonomous driving in general, and autonomous racing in specific, replace partial or whole modules of the modular perception-planning-control autonomous software pipeline with data-driven approaches\cite{betz2022autonomous}. For instance, \cite{Tatulea2020NMPCDNN} combined non-linear model predictive
control (NMPC) and deep neural network (DNN) for the trajectory planning, while \cite{Chisari2020SAC} utilize model-based RL to test their vision-based planning. 


Few studies have explored IL for autonomous racing. Deep imitation learning (DIL) \cite{Pan2018DIL} trained the DNN policy with an MPC expert using DAGGER and tested it on AutoRally. Additionally, few studies took IL and RL together into account. Controllable imitative reinforcement learning (CIRL) \cite{Liang2018ECCV}, and deep imitative RL (DIRL) \cite{Cai2021DIRL} initialized the RL policy network with IL before starting exploration. Still, IL was implemented in basic behavioral cloning (BC) rather than interactive IL methods. \cite{Zou2020DDPGIL} and \cite{Vecer2017DDRL} loaded the transitions of demonstration into the replay buffer
to lead the RL process \cite{Cai2021DIRL}. Nonetheless, they still use IL as simple demonstrations, and the methods have not been verified in real-world scenarios. 

\subsection{Imitation Learning}
BC is the most straightforward IL method. Supervised machine learning is used in BC to train the novice policy with the demonstrated expert policy. One of its first applications in autonomous driving from 1988 is ALVINN \cite{ALVINN}, which could achieve the vehicle-following task on the road with a vehicle equipped with sensors. BC is easy to understand and simple to implement. However, it suffers from the risk of distribution mismatch, covariate shift \cite{ARGALL2009survey}, and compounding errors \cite{ross2011dagger}, making it brittle for autonomous racing.

Studies in imitation learning after BC usually could be categorized into direct policy learning (DPL) and inverse reinforcement learning (IRL). DPL primarily emphasizes learning the policy directly \cite{ahmed2017survey}, while IRL pays more attention to learning the intrinsic reward function \cite{Arora2021IRL}. In this paper, we will mainly discuss DPL.

Data Aggregation (DAGGER) allows the novice to influence the sampling distribution by aggregating the expert-labeled data extensively and updating the policy iteratively, which mitigates BC’s drawbacks \cite{ross2011dagger}. However, the data-gathering rollouts are under the complete control of the incompletely trained novice rather than interactively querying the expert, which degrades the quality of the sampling and efficiency of data labeling; this even potentially destabilizes the autonomous system \cite{kelly2019hgdagger}.

To deal with the drawbacks of DAGGER, recent developments in interactive IL involved the human-gated method and robot-gated method. The human-gated techniques, e.g., \cite{kelly2019hgdagger,Spencer2020LfI}, allow the human supervisor to decide the instants to correct the actions, but continuously monitoring the robot will burden the supervisor. The robot-gated method \cite{Zhang2016SafeDAgger} enables the robot to query the human expert for interventions actively but balancing the burden and providing sufficient information is still difficult \cite{triftydagger}.
In this paper, we only consider the human-gated method since the robot-gated method is unsuitable for our racing conditions with low reaction time. Using the robot-gated method will probably cause the crashing due to high-speed racing and inertia.

By introducing the gating function, human-gated DAGGER \cite{kelly2019hgdagger} allows the human expert to take control when the condition is beyond the safety threshold and to return the control to novice policy under tolerated circumstances. The interactive property reduces the burden of the expert rather than querying the expert all the time and ensures the efficiency of the training process. 
Expert intervention learning (EIL) \cite{spencer2022eil} proposed further exploration with implicit and explicit feedback beyond HG-DAGGER. It addresses that any amount of expert feedback needs to be considered, whether intervened or not. EIL records the data into three state-actions datasets and added the implicit loss inferred when the expert decides to intervene.

\begin{table*}[t]
\small
\caption{Evaluations of different learned policies in an unseen simulation environment.}
\label{eval_generalizability}
\setlength{\tabcolsep}{3.3pt}
\begin{tabular}{*{10}{c}}
\toprule
Method  & BC & DAGGER & HG-DAGGER & EIL & PPO & BC+PPO & DAGGER+PPO & HG-DAGGER+PPO & EIL+PPO \\
\midrule
Distance Traveled (\si{m}) & 7.84 & 8.90 & 12.34 &  15.89 & 12.69 & 151.23 & 86.49 & 155.88 & 150.15 \\
Complete 1 Lap & No & No & No & No & No & Yes & No & Yes & Yes \\
Bhattacharyya Distance & 0.77 & 0.60 & 0.12 & 0.24 & 1.09 & 0.59 & 0.59 & 0.47 & 0.43 \\
\bottomrule
\end{tabular}
\vspace{-0.3cm}
\end{table*}

\section{METHODOLOGY}


The IL algorithms we implement and compare in this work include BC \cite{sammut2010bc}, DAGGER \cite{ross2011dagger}, HG-DAGGER \cite{kelly2019hgdagger}, and EIL \cite{spencer2022eil}. We use a multi-layer perceptron (MLP) as the learner and a pure pursuit algorithm as the expert for all implemented IL algorithms. We are using a 2D simulation environment (F1TENTH gym) developed for the small-scale autonomous race car \cite{Betzf1tenth2022}. The learner takes the LiDAR scan array \(o_{L}\) as input, whereas the pure pursuit expert takes the \(x\) and \(y\) coordinates and angular direction of the agent \(\theta\) as input. Both the learner and expert output steering angle and speed as actions to control the vehicle.

For human-gated IL algorithms that require expert intervention, which are HG-DAGGER and EIL, we define two intervention thresholds \(\gamma_{v}\) and \(\gamma_{\omega}\) for speed \(v\) and steering angle \(\omega\), respectively. To mimic the expert intervention using the pure pursuit algorithm, both the learner policy and the pure pursuit output their action based on their observation separately at every step. The pure pursuit will take over the control and provides expert demonstrations whenever the difference of action between the learner policy and the pure pursuit exceeds either of \(\gamma_{v}\) or \(\gamma_{\omega}\). 

Considering the prominence of RL in learning-based methods for autonomous racing and the potential of IL as a bootstrapping method for RL, we also implement proximal policy optimization (PPO) \cite{schulman2017ppo} to train policies with or without IL bootstrapping to compare the efficiency of various combinations of PPO and different IL algorithms. Before training, the PPO policy can be initialized randomly or bootstrapped using a pre-trained network by IL algorithms with \(n\) expert-labeled samples. The pre-trained IL network has the same architecture with the PPO policy. To reduce warbling, avoid crashing and encourage staying close to the center-line of the track, we design a reward function \(r\) that incorporates the reward for survival, the penalty for lateral error from the center-line \(E_l\), and penalty for the deviation of steering angle \(\omega\) as
$$
r = -0.02 \cdot \min(1.0, \max(0, \omega)) + \begin{cases}
                                              -0.5 & \text{if crashed}\\
                                              -0.02 \cdot E_{l}  & \text{if \(E_{l} > 0.1\)}\\
                                              0.02 & \text{otherwise}
                                            \end{cases}   
$$

To transfer the learned policy from simulation to the real world, we add an array of random noise \(o_R\) to the LiDAR scan array \(o_{L}\). \(o_R\) and \(o_L\) have the same dimension. Each element in \(o_R\) is randomly sampled from \([\alpha, \beta]\), where \(\alpha\) and \(\beta\) are the lower and upper bounds of the random noise.

\vspace{-0.12cm}
\begin{figure}[h]
      \centering
      \includegraphics[scale=0.39]{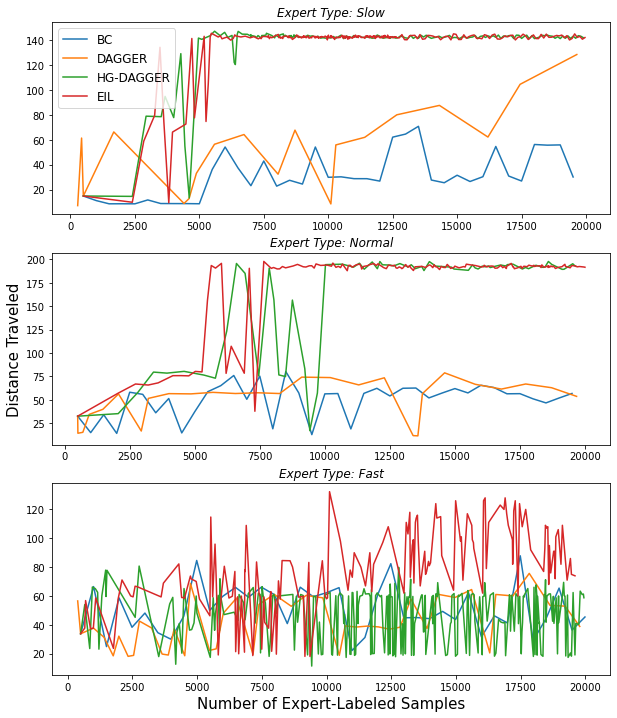}
      \caption{The distance traveled by each agent with respect to the number of expert-labeled samples.}
      \label{sim_distance_traveled}
\end{figure}

\section{EXPERIMENTS}
\subsection{Implementation Details}
We implement the IL algorithms using the F1TENTH, a 1/10-scale autonomous racing research platform, for both simulation and real-world scenarios\cite{Betzf1tenth2022}. The maps for training and evaluation in simulation and the real world are shown in Fig. \ref{maps}. Due to safety considerations, all policies in this work are trained in the F1TENTH gym. We use the same two-layer MLP with 256 hidden units as the learner for all IL algorithms in the comparison. The learning rate is set to 0.001 during training. When training policies using DAGGER, HG-DAGGER, and EIL, the first 500 samples are collected for training initial policies using BC. For HG-DAGGER and EIL, we set the intervention threshold \(\gamma_{v}\) and \(\gamma_{\omega}\) to 1 and 0.1, respectively. All IL policies are trained using 20k expert-labeled samples. To test the efficiency of bootstrapping, we train different PPO policies for 20k steps. We use IL policies with 3000 expert-labeled samples as the starting point for bootstrapped PPO policies. We let $\alpha= -0.2$ and $\beta= 0.2$ for transferring policies to real world. 


\subsection{Evaluations in Simulations}

\begin{table}[t]
\small
\caption{Elapsed time of IL policies trained with different experts}
\label{elapsed_time}
\setlength{\tabcolsep}{1.8pt}
\begin{tabular}{*{6}{c}}
\toprule
 Expert Type & Expert & BC & DAGGER & HG-DAGGER & EIL\\
\midrule
Slow & 33.07 \si{s} & Failed & 34.34 \si{s} & 33.78 \si{s} &  33.50 \si{s} \\
Normal & 25.04 \si{s} & 25.35 \si{s} & 25.85 \si{s} & 25.06 \si{s} & 25.22 \si{s} \\
Fast & 19.69 \si{s} & Failed & Failed & Failed & 20.40 \si{s} \\
\bottomrule
\end{tabular}
\end{table}

\begin{table*}[t]
\small
\caption{Evaluations of different learned policies in the real-world environment.}
\label{real_world_table}
\setlength{\tabcolsep}{2.65pt}
\begin{tabular}{*{11}{c}}
\toprule
Method  & Expert & BC & DAGGER & HG-DAGGER & EIL & PPO & BC+PPO & DAGGER+PPO & HG-DAGGER+PPO & EIL+PPO \\
\midrule
Distance Traveled (\si{m}) & 61.44 & 6.44 & 8.49 & 37.74 &  38.04 & 5.27 & 6.44 & 6.29 & 64.08 & 39.5 \\
Complete 1 Lap & Yes & No & No & No & No & No & No & No & Yes & No \\
\bottomrule
\end{tabular}
\end{table*}

\begin{figure*}[h]
      \centering
      \includegraphics[scale=0.266]{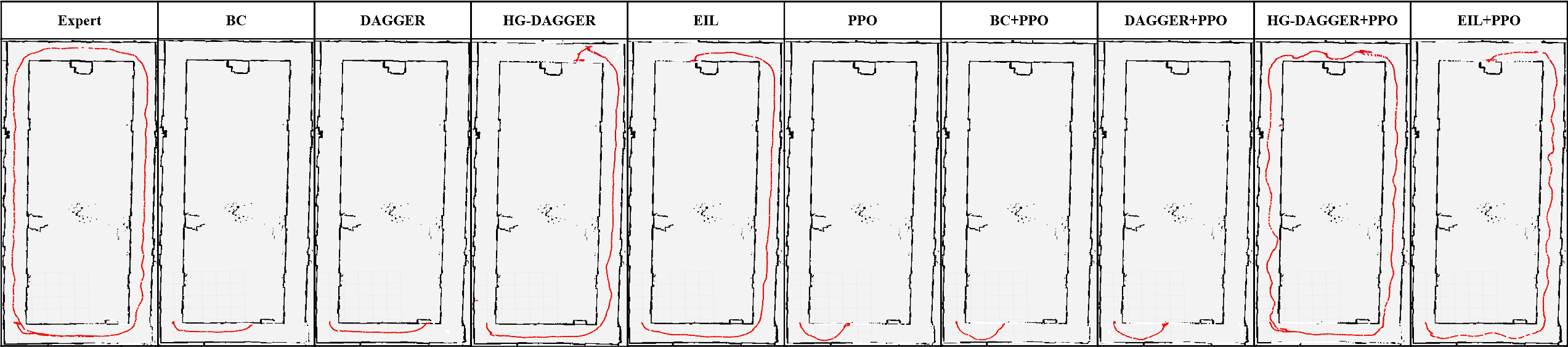}
      \caption{The movement trajectories of the F1TENTH vehicle on the map under the control of each policy.}
      \label{real_world_trajs}
      \vspace{-0.3cm}
\end{figure*}

During the training of the four IL algorithms, the learned policies are evaluated in terms of distance traveled in the training map after the training of each iteration. We use the number of expert-labeled samples as the independent variable to assess and compare the sample efficiency of different IL algorithms for three reasons. First, the major downside of IL, in general, is its requirement of expert effort with labeling. Moreover, the number of steps in each iteration is not fixed and is uncontrollable for all algorithms except BC. Lastly, implicit samples in EIL are collected at no cost, which makes it unfair to compare with other algorithms in terms of the total number of samples.

\begin{figure}[h]
      \centering
      \includegraphics[scale=0.39]{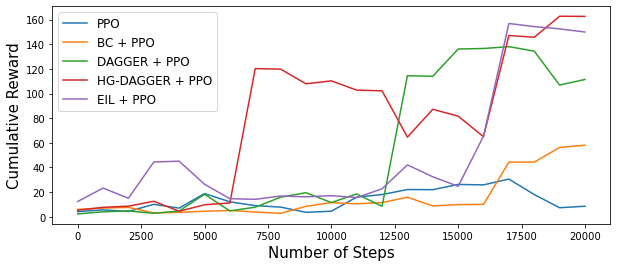}
      \caption{The cumulative reward with respect to the number of steps during training PPO policies with or without IL bootstrapping.}
      \label{sim_bootstrap}
      \vspace{-0.3cm}
\end{figure}

As shown in Fig. \ref{sim_distance_traveled}, to further test the ability to imitate the expert's behavior for each IL algorithm, we train different policies using the demonstration of the pure pursuit expert with different speeds. The slow, normal, and fast experts are with an average speed of 4.79 \si{m/s}, 6.39 \si{m/s}, and 8.24 \si{m/s} respectively. Overall, those IL algorithms with expert intervention, i.e., HG-DAGGER and EIL, have better sample efficiency since their learned policies travel significantly longer distances than BC and DAGGER. 

Although all IL algorithms struggle to learn when using the demonstrations from the fast expert, as shown in Table \ref{elapsed_time}, EIL is the only one that can complete one lap. Table \ref{elapsed_time} also indicates that the upper limit of the performance of policies learned using IL is from the expert.




Since IL can be combined with RL, we test the bootstrapping efficiency of different IL algorithms for PPO at normal speed. As Fig. \ref{sim_bootstrap} suggests, using any IL algorithms can help PPO converge to a better policy as it no longer needs to start from a random policy. DAGGER, HG-DAGGER, and EIL demonstrate considerably better bootstrapping efficiency than BC, thanks to their better sample efficiency.

To evaluate how well the policies generalize, we generate a new (unseen) map, as shown in Fig. \ref{maps}(b), and perform inference using the policies at normal speed. For each policy, we record the distance traveled and whether it completes one lap or not to evaluate its performance. We also calculate the Bhattacharyya distance \cite{fukunaga1990feature} for the decision of an expert at every step to evaluate the similarity between learned behaviors and expert behavior. As Table \ref{eval_generalizability} shows, the combinations of IL and PPO significantly outperform policies only using IL or PPO, with the PPO trained with EIL bootstrapping having the best performance. This indicates that combining IL and RL can efficiently converge to a more generalized policy. Additionally, interactive IL can train policies that are more similar to expert behavior than non-interactive IL and PPO.

\subsection{Evaluations in Real-World Environments}
All policies for real-world experiments are at 3 \si{m/s}. Fig. \ref{real_world_trajs} shows the results of real-world experiments. For both direct training and bootstrapping, interactive IL methods, which are HG-DAGGER and EIL, can train policies that travel considerably further distances compared with policies from non-interactive methods, which are BC and DAGGER. BC and DAGGER barely help when bootstrapping PPO in real-world experiments. The combination of HG-DAGGER and PPO has the best performance and is the only policy that completes one lap in the real world. Despite incorporating the penalty on steering angle in the reward function, PPO policies have more warbling in their trajectories compared with IL policies, which might result from the difference in floor friction between the gym environment and the real world. The real-world experiments further validate that combining IL and RL yields the best result.

\section{CONCLUSION}

In this work, we implement four different IL algorithms on the F1TENTH platform to benchmark their performance in the context of autonomous racing. Our experiments show that IL algorithms can train or bootstrap high-performance policies for autonomous racing scenarios. Recent development in interactive IL significantly improves the sample efficiency of policies for autonomous racing. The combination of RL and interactive IL can get the best of both worlds: fast convergence and better generalizability. The interactive imitation learning methods outperform non-interactive methods for both learning directly and bootstrapping due to their improved sample efficiency. Our IL implementations provide a foundation for future research on autonomous racing using IL and RL. Future work will focus on safe human-gated methods for multi-agent autonomous racing, utilization of new network architecture, and better simulation environments to further reduce the sim-to-real gap.

\section{Acknowledgment}
This work was supported in part by NSF CCRI 1925587 and DARPA FA8750-20-C-0542 (Systemic Generative Engineering). The views, opinions, and/or findings expressed are those of the author(s) and should not be interpreted as representing the official views or policies of the Department of Defense or the U.S. Government.

\bibliography{references.bib}

\end{document}